\documentclass{article}

\usepackage[english]{babel}

\usepackage[letterpaper,top=2cm,bottom=2cm,left=3cm,right=3cm,marginparwidth=1.75cm]{geometry}

\usepackage{amsmath}
\usepackage{graphicx}
\usepackage[colorlinks=true, allcolors=blue]{hyperref}

\usepackage{xcolor}         %
\PassOptionsToPackage{numbers, compress}{natbib}
\usepackage{natbib}

\usepackage{url}            %
\usepackage{booktabs}       %
\usepackage{amsfonts}       %
\usepackage{nicefrac}       %
\usepackage{microtype}      %
\usepackage{graphicx}
\usepackage{enumitem}
\usepackage{amsmath}
\usepackage{pifont}
\usepackage{placeins}
\usepackage{braket}
\usepackage{soul}
\usepackage{subcaption}
\usepackage{bm, bbm}

\usepackage{algorithm, algorithmic}

\title{\bf The Need for Verification in \\AI-Driven Scientific Discovery}

\author{
 Cristina Cornelio\thanks{Samsung AI, Cambridge, UK {\tt\small c.cornelio@samsung.com}} 
 ~~~~Takuya Ito\thanks{IBM Research, Yorktown Heights, USA}
 ~~~~Ryan Cory-Wright\thanks{Imperial Business School, London, UK}  \\
 Sanjeeb Dash\footnotemark[2]
 ~~~~Lior Horesh\footnotemark[2]
}

\date{}

\begin{document}
\maketitle
\begin{abstract}
Artificial intelligence (AI) is transforming the practice of science. Machine learning and large language models (LLMs) can generate hypotheses at a scale and speed far exceeding traditional methods, offering the potential to accelerate discovery across diverse fields. However, the abundance of hypotheses introduces a critical challenge: without scalable and reliable mechanisms for verification, scientific progress risks being hindered rather than advanced. In this article, we trace the historical development of scientific discovery, examine how AI is reshaping established practices for discovery, and review the principal approaches, ranging from data-driven methods and knowledge-aware neural architectures to symbolic reasoning frameworks and LLM agents. While these systems can uncover patterns and propose candidate laws, their scientific value ultimately depends on rigorous and transparent verification, which we argue must be the cornerstone of AI-assisted discovery.
\end{abstract}

\section{Introduction}

The overarching goal of science is to provide a set of universal, accurate, and interpretable explanations that describe the natural world. This involves discovering natural laws that {not only make accurate predictions but are also corroborated by scientific experiments and existing scientific literature}. Such laws have historically been discovered through the scientific method, a systematic process that begins with a question and proceeds through a study phase during which researchers gather all prior knowledge {and data} pertaining to the phenomenon under investigation. This leads to the formulation of hypotheses, empirical validation, and \noindent
iterative refinement. The scientific method enables the discovery of verifiable scientific truths by relying on substantiated and repeatable evidence, lending science its legitimacy and credibility.

The shift from dogmatic belief systems to a framework grounded in scientific theory and empirical verification, epitomized by the 16th-century transition from religious authority to human reason through the work of Copernicus, Galileo, and Bruno, marked a fundamental epistemological transformation leading to the Scientific Revolution~\citep{leveillee2011copernicus}.  
Indeed, Kepler's mathematical description of planetary motion, grounded in Tycho Brahe’s observational data and Bacon’s advocacy for inductive reasoning, established the foundation for modern empirical science~\citep{bacon1620novum, westfall1971construction}. 
Critically, the consequences of this verification-driven methodology have been profound in the modern age.
{Throughout the text, we use {\em verification} to mean validating a hypothesis against scientific knowledge, whether based on theoretical axioms or empirical data, and we use {\em hypothesis}  to mean a testable and falsifiable statement about the systematic relationship between two or more variables of interest~\citep{popper_logic_1959}.}

{Empirically} verified discoveries such as germ theory~\citep{latour1988pasteurization}, high-yield agricultural practices~\citep{borlaug1970nobel}, and thermodynamic principles~\citep{smil2001enriching} have transformed medicine, food production, and energy systems. These advances were achieved through a disciplined integration of theoretical models and experimental validation.

\begin{figure}[h]
    \centering
    \includegraphics[scale=1.0]{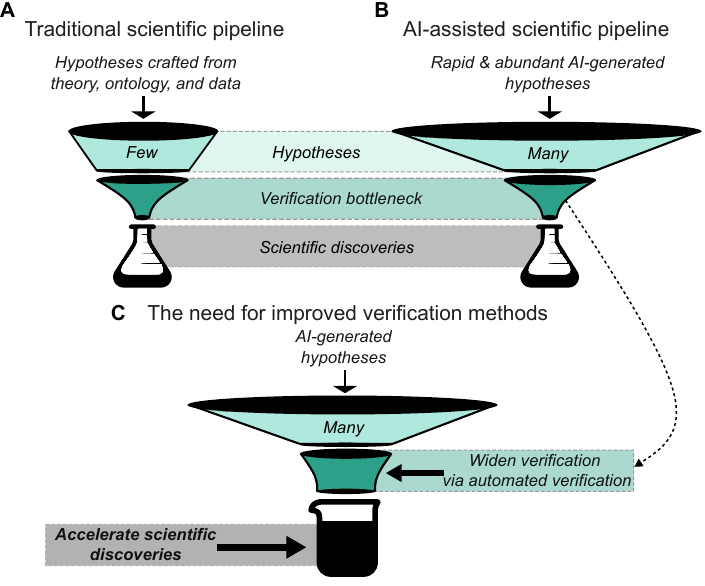}
    \caption{\textbf{The scientific method in the age of AI.} 
    A) In the traditional scientific method, theories guide the generation of testable hypotheses, which are then validated through experiments and data.
    B) However, with generative AI, hypotheses can be rapidly produced from data, but {formal logic} verification still relies on slow, manual evaluation by domain experts.
    C) Without widening this verification bottleneck (e.g., through automated/integrated verification){,} the pace of discovery remains limited, despite the acceleration promised by AI. %
    }
    \label{fig:hypothesis_generation}
\end{figure}

However, the rate of major discoveries has declined in both absolute and relative terms over the past several decades~\citep{bloom2020ideas, bhattacharya2020stagnation, arora2018decline}. This decline is arguably due to the exhaustion of simple, low-dimensional theories, and the increasing complexity of modern scientific problems~\citep{cowen2011great} (see Fig.~\ref{fig:hypothesis_generation}A).
Amid these challenges, the rise of machine learning and artificial intelligence has introduced promising tools to augment hypothesis generation and data analysis for scientific discovery.
However, many existing implementations of these data-driven systems lack formal mechanisms for logical inference that are essential for {logically} verifiable scientific discovery~\citep{platt_strong_1964}.
In particular, generative AI models have shown remarkable capacity to rapidly generate novel scientific hypotheses~\citep{gottweis2025towards,yamada2025ai,lu2024ai,jumper_highly_2021}.
However, these outputs often lack empirical grounding and are frequently disconnected from established theoretical frameworks or domain-specific knowledge. 
This disconnect has led to an overwhelming influx of hypotheses, straining verification pipelines that are essential for validating scientific discoveries~\citep{beel2025evaluating,gridach2025agentic,kulkarni2025scientific} (see Fig. \ref{fig:hypothesis_generation}B). 
Consequently, developing robust methods to refine and verify hypotheses from data-driven approaches is critical to unlocking the full potential of AI in accelerating scientific progress (Fig. \ref{fig:hypothesis_generation}C).

To address the limitations of purely data-driven approaches, several recent works propose hybrid frameworks.
These approaches integrate machine learning with elements of symbolic reasoning, constraint imposition, and formal logic, aiming to ensure scientific validity alongside predictive accuracy (e.g., see~\cite{wiberg2025synergizing} for a review {of the} integration between AI/ML and Operations Research techniques). 
For example, Kolmogorov–Arnold Networks (KANs)~\citep{liu2024kan} replace fixed linear weights with learnable univariate functions, producing interpretable approximations of scientific relations. 
Hamiltonian Neural Networks (HNNs)~\citep{greydanus2019hamiltonian} similarly enforce energy conservation by learning a Hamiltonian and deriving system dynamics from Hamilton’s equations. While the learned models from both systems respect structural embeddings, verification for both systems is limited to their specific structural propert{ies}. 
More recently, AI-Descartes~\citep{cornelio_combining_2023} introduced a general verification mechanism {based on logic derivability}, where hypotheses were generated via a data-driven approach and later verified against known theory via theorem proving. 
Building upon that work, AI-Hilbert~\citep{cory2024evolving} integrated data and theory directly during the hypothesis generation stage, thereby constraining the search to expressions consistent with both data and theory. While both these approaches provide scientifically verifiable results {(via logic derivability)}, their application is limited to specific problem formulations in the physical sciences.
Thus, while emerging computational tools offer great promise for broadly accelerating scientific discovery, their effectiveness hinges on ensuring {the} resulting insights are not only predictive but also interpretable, {logically derivable}, and aligned with foundational scientific knowledge.

In this article, we review recent progress in AI-driven scientific discovery, while underscoring the critical importance of verifying these methods throughout the discovery process.
We begin by highlighting key historical examples where the failure to rigorously verify computational methods led to {serious scientific errors and wasted resources}. Next, we examine recent data-driven approaches to scientific discovery, highlighting their ability to uncover patterns and generate hypotheses from large datasets, particularly in domains where theoretical models are incomplete or unavailable. 
This is followed by a comparison with knowledge-aware methods, the emergence of derivable models that integrate symbolic reasoning, and the growing role of large language models (LLMs) in automating and augmenting scientific workflows.
We conclude with a broader perspective on the heterogeneous role of verification across scientific domains, outlining current challenges and suggesting promising approaches for future research.

\section{The Importance of Verification: Evidence from Examples}

{Verification is the foundation that ensures scientific discoveries are real, reproducible, and trustworthy. It is essential for both paradigm-shifting discoveries~\citep{kuhn2012structure} and incremental advances. 
For revolutionary findings, it rarely depends on a single experiment or theory.
Instead, it demands converging evidence -- from rigorous data validation and independent replication to theoretical coherence. 
This same principle applies to incremental science, where careful verification ensures reliability, replicability, and progress.
In 1924, astronomer Walter Sydney Adams claimed to have confirmed the gravitational redshift predicted by general relativity, but both his measurements and theoretical assumptions were flawed. 
For decades, critics cited this as evidence against Einstein’s theory until later experiments corrected the mistake \citep{adams1925relativity}.
In molecular biology, early claims of arsenic-based life forms were later invalidated, {and the inability to replicate empirical results} led to years of wasted scientific resources \citep{wolfe2011retracted}. 
In chemistry, minor incorrect stoichiometric calculations during synthesis have led to widely cited but ultimately retracted papers, such as reports of novel catalytic activity that could not be reproduced because of miscalculated reagent ratios \citep{xu2021retracted}.
The lesson from these domains is clear: without rigorous verification, minor errors can compound, leading to wasted resources and efforts, independent of the scale or scope of the scientific inquiry.}

In automated scientific discovery, the same principle applies. {Automated model-generation tools have begun to show impact in narrow domains, illustrating their potential to transform scientific discovery.}
Symbolic regression engines such as PySR~\citep{pysr} and AI Feynman~\citep{udrescu2020ai}, {Bayesian-based methods~\citep{guimera2020bayesian},} as well as neural architectures like Kolmogorov–Arnold Networks (KANs)~\citep{liu2024kan}, Hamiltonian Neural Networks (HNNs)~\citep{greydanus2019hamiltonian}, and Lagrangian Neural Networks (LNNs)~\citep{cranmer2020lagrangian}, now produce many new hypotheses quickly. This proliferation creates a new bottleneck: distinguishing between formulas that merely fit the data and those that are scientifically meaningful (Fig. \ref{fig:hypothesis_generation}). Without rigorous verification {(e.g., checking logical derivability or consistency from a background theory, verifying constraints, or empirically validating on numerical/experimental data)}, the flood of generated hypotheses risks overwhelming the scientific process with plausible but superficial results. Thus, verification is an essential filter that separates genuine scientific discoveries from hallucinations or mere noisy interpolations that fail to generalize beyond the observed data.

The challenge of verification is further exacerbated by the rise of LLM-based tools whose reliability can be strongly questioned~\citep{marcus2025blogpost,kambhampati2024reason}. 
Well-publicized examples of hallucinations from LLMs include the hallucination of legal cases cited in court filings~\citep{weiss2023lawyer}, the fabrication of biomedical references~\citep{gravel2023learning}, and outputs violating basic algebraic~\citep{hendrycks2021math} or physical consistency~\citep{wang2023newton}. 
We refer to~\cite{zhang2025exploring} for a recent review of LLMs and their use in scientific discovery. 

One might argue that the latest generation of large language models (LLMs), which utilize Reinforcement Learning (RL), can be trained and steered to produce meaningful responses.
In particular, RL with Verified Rewards (RLVR) trains models to produce so-called reasoning chains that lead to verified responses \citep{shao2024deepseekmath}.
However, in these settings, verification of the response is not equivalent to rigorous logical verification for several reasons. First, RLVR merely verifies the response rather than the intermediate steps (or chain of thought) used to arrive at it. Moreover, models have been shown to {\em produce incorrect reasoning steps to arrive at correct responses} \citep{kambhampati2025stop,stechly2025beyond}. Second, some studies have shown that LLMs can improve benchmark performance on RLVR datasets that use randomized and even negative rewards for correct responses~\citep{shao2025spurious}. Third, RLVR provides no guarantees; performance gains from RLVR are idiosyncratic with the base pretrained model used~\citep{shao2025spurious}. Nevertheless, these findings do not preclude RL from being a useful tool when combined with other tools. For example, AlphaGeometry, AlphaEvolve, and their related descendants have been successfully used in scientific discovery, particularly when combined with other symbolic methods \citep{novikov2025alphaevolve,chervonyi2025gold,zhang2024proposing}. {At the same time,} while applying RL to LLMs can improve benchmark performances and the style and surface reliability of generated responses (e.g., through RL with Human Feedback~\citet{ziegler2019fine}), it does not address the deeper need for principled, automated logic verification against background theory (e.g., derivability or consistency checking) and empirical constraints.
For scientific discovery, this distinction is crucial: plausibility without proof cannot serve as the foundation of knowledge.

In the mathematical literature, the use of formal proof assistants for verification, such as Lean \citep{de2015lean}, Coq \citep{bertot2013interactive}, and Isabelle \citep{nipkow2002isabelle}, has attracted considerable attention. These systems enable mathematical theorems to be expressed in a dependently typed language and logically verified computationally.
For example, they can be used to verify that every result in an introductory analysis textbook is correct~\citep{tao2025blogpost}. Moreover, some recent works pair this technology with LLMs. For instance, the AlphaProof system from DeepMind~\citep{deepmind2024blogpost} achieved a silver medal at the 2024 International Mathematics Olympiad by translating one million informal mathematical problems into Lean using natural language processing, allowing AlphaProof to be trained using reinforcement learning. However, since there is no commonly agreed-upon set of axioms for the natural sciences (e.g., quantum mechanics and gravity are not consistent), and the process of translating informal problems into formal statements can introduce errors unless verified by a user, a Lean-reinforcement learning approach cannot be broadly applied to scientific discovery.

Finally, the growing recognition of verification challenges in AI-driven scientific discovery has catalyzed significant government investment in bridging formal methods with statistical AI approaches. 
In the United States, DARPA’s portfolio is an example of this trend, featuring programs such as \textit{expMath}~\citep{darpa-expmath}, which seeks to accelerate mathematics by developing AI systems capable of proposing and proving abstractions, and \textit{The Right Space (TRS)}~\citep{darpa-trs}, which applies scientific machine learning to uncover tractable transformations for complex models. Other initiatives~\citep{darpa-remath, darpa-provers, darpa-vspells, darpa-hacms}, including \textit{ReMath}, \textit{PROVERS}, \textit{V-SPELLS}, and the completed \textit{HACMS} program, further underscore the emphasis on formal verification in critical domains.
These funding priorities reflect an acknowledgment that while generative AI excels at quick hypothesis generation and pattern discovery, scientific applications require the reliability and guarantees that only formal verification methods {(such as logic derivability)} can provide.

\section{AI Methods for Scientific Discovery}

\begin{figure}[h]
\centering
\includegraphics[width=1.0\textwidth]{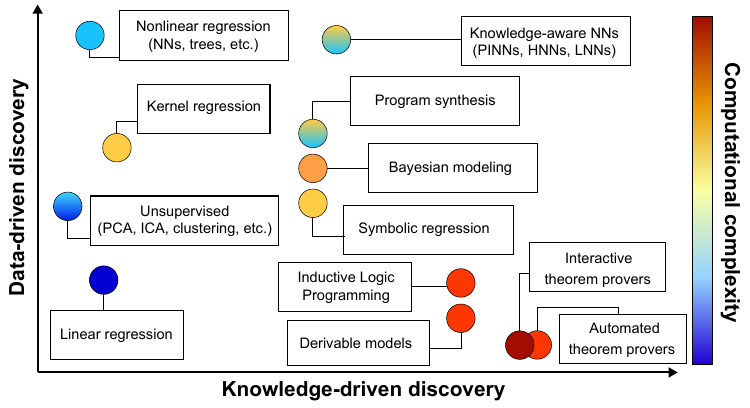}
\caption{\textbf{Qualitative Landscape of Computational Methods for Scientific Discovery.} Different approaches span the spectrum between data-driven and knowledge-driven discovery. Data-driven methods, such as neural networks, can rapidly generate hypotheses but lack verifiability, whereas theory-driven methods, like automated theorem provers, offer rigorous verification {(logic derivation)} but are often slow and undecidable. Derivable or science-aware approaches aim to bridge this gap by combining data-driven modeling with symbolic guarantees. The associated computational complexity reflects trade-offs between speed, interpretability, and verifiability.
Note that this figure provides a qualitative illustration of the landscape of computational tools for scientific discovery, highlighting general trends across major categories. 
The position of specific methods in these categories may vary depending on data type, approach, or hybrid usage.
}
\label{fig:datavknowledge1}
\end{figure}

Motivated by the increasing importance of verification in scientific discovery and other domains involving AI, we next review the state-of-the-art methods used for scientific discovery. Figure~\ref{fig:datavknowledge1} provides a qualitative map of the landscape, positioning the different {categories of methods} along three dimensions: the degree to which they are data-driven, the degree to which they are knowledge-driven{,} and their associated computational complexity. We also discuss the limitations of the existing approaches and suggest strategies for future improvement.

\subsection{Data{-}Driven Methods}

The data-driven discovery of symbolic formulae is a long-standing challenge in artificial intelligence~\citep{Kitano_2016}, and the central difficulty remains how to incorporate verification into the process.
{Building upon centuries of experience building models using small datasets~\citep{russell1964kepler}, a}
variety of {big-data} approaches have been proposed~\citep{landajuela2022equation}, ranging from neural networks designed to mimic human physical reasoning~\citep{NN_discovery}, to tree-structured LSTMs for handling symbolic expression trees and formula verification~\citep{symb-and-neural}, to logic-constrained GANs for image generation~\citep{visual_learning_logic}. Symbolic regression (SR) has played a prominent role in this space, with applications to extracting explicit relations from graph neural networks~\citep{cranmer2020discovering}, constructing analytic models for reinforcement learning control~\citep{symb-reg-analytic}, or combining regression with Bayesian models~\citep{bayesianSR}. The AI Feynman family of methods~\citep{udrescu2020ai,AI-Feynman2.0} exemplifies the integration of neural fitting with physics-inspired heuristics, while tools such as PySR~\citep{pysr, cranmer2020discovering} and TuringBot~\citep{schmidt2009distil} use evolutionary or annealing-based search strategies to identify parsimonious equations. More recent methods, such as RSRM, combine Monte Carlo Tree Search with reinforcement learning for efficient symbolic exploration~\citep{wang2024rsrm}. Despite the progress, none of these approaches incorporates formal reasoning, leaving their outputs vulnerable to producing expressions that fit the data but lack theoretical grounding.

To address this gap, several works have attempted to combine SR or neural methods with logical consistency checking. The LGML system~\citep{scott2021lgml} augments learning with a module that verifies whether candidate functions satisfy constraints on their functional form, while LGGA~\citep{ashok2020logic} extends this approach with genetic algorithms and auxiliary mathematical expressions. Similar ideas appear in \cite{bladek} counterexample-guided SR, in \cite{kubalik2020symbolic,kubalik2021multi} multi-objective framework that enforces nonlinear constraints as discrete data points, and in \cite{engle2021deterministic} deterministic mixed-integer programming formulation with derivative constraints. These methods, however, remain limited to constraints on functional form rather than incorporating background-theory axioms that describe the scientific environment itself.

In parallel, the broader neuro-symbolic community has explored the integration of logic constraints into machine learning tools. Approaches include penalizing constraint violations in neural networks~\citep{xu2018semantic,wang2020integrating} and embedding logical rules in the training process~\citep{cornelio_2023_NASR,li2019augmenting,daniele2020neural,xie2019embedding,li2019logic}. Inductive logic programming and rule induction~\citep{tamaddoni2021human,Riegel_lnn,evans2018learning,NEURIPS2019_0c72cb7e,law2018inductive} provide another way of extracting logical knowledge from data, while program synthesis has gained renewed interest as a means of combining symbolic reasoning with statistical learning~\citep{sun2022neurosymbolic, NEURIPS2020_7a685d9e,parisotto2016neuro,valkov2018houdini,yang2017differentiable}. Yet, across all these efforts, formal verification {(e.g., logic derivability)} of discovered formulas remains elusive: constraints typically ensure plausibility, not provability. The result is that many systems generate equations that appear valid but are not guaranteed to align with the underlying laws of nature.

\subsection{Knowledge-Aware Methods}

Scientific discovery and artificial intelligence have traditionally followed separate paradigms: the former rooted in theory and verification, and the latter in data-driven learning. As scientific problems become increasingly complex and data become increasingly abundant yet noisy or incomplete, there is a growing interest in integrating scientific knowledge into machine learning models. The resulting hybrid methods aim to combine the flexibility of learning-based approaches with the structure and generalizability offered by physical laws.

This section surveys approaches that leverage scientific knowledge in AI model design and training. 
We distinguish between \textit{physics-informed} models, which learn the unknown solution of known governing equations by training neural networks to minimize data and physics residuals, and \textit{physics-inspired} models that encode known structures, such as conservation laws, directly in the network's architecture. We also consider symmetry-informed networks that embed invariance or equivariance directly into model operations, so that transformations of the input induce consistent transformations of the output.

\paragraph{Physics-Informed Neural Networks (PINNs).}

PINNs incorporate governing physical laws into the learning process by embedding partial differential equations (PDEs) directly into the loss function~\citep{raissi2019physics, cuomo2022scientific}. These models are not intended to discover the governing equations themselves, but rather to approximate their solutions. The key idea is to replace or augment traditional numerical solvers by training a neural network that minimizes a composite loss consisting of:
1) A data loss term measuring the fit to observed data;
2) A physics loss term penalizing violation of the PDE; and
3) A boundary condition loss term ensuring physical consistency.
Mathematically, for a PDE of the form $F(x, u, \nabla u, \nabla^2 u) = 0$, the PINN approximates the solution $u(x)$ with a neural network $u_\theta(x)$, and minimizes:
\[
\mathcal{L}_{\text{total}} = \lambda_d \mathcal{L}_{\text{data}} + \lambda_f \mathcal{L}_{\text{physics}} + \lambda_b \mathcal{L}_{\text{boundary}}.
\]

This approach has demonstrated success across various domains, including fluid mechanics, heat diffusion, and quantum mechanics. It provides an elegant, mesh-free framework capable of solving high-dimensional PDEs with limited data.

While PINNs represent a significant advance in scientific computing, the method requires carefully balancing multiple loss terms, and is therefore sensitive to network architecture choices.
This highlights the importance of systematic hyperparameter optimization strategies. Additionally, the current approach relies on known governing equations as constraints, and the generic network architectures employed do not yet fully exploit problem-specific structural information. These characteristics have motivated active research directions focused on adaptive loss weighting schemes, physics-informed architecture design, and methods for discovering unknown governing equations from data~\citep{lu2021deeponet}.

\paragraph{Physics-inspired Neural Networks.}

Physics-inspired neural networks take a complementary approach: instead of embedding the governing equations into the training loss, they encode physical structure directly into the model architecture. These models are well-suited to systems governed by conservation laws, such as those following Hamiltonian or Lagrangian dynamics.

In \textit{Hamiltonian neural networks (HNNs)}~\citep{greydanus2019hamiltonian}, the model learns a scalar-valued Hamiltonian function $H(q, p)$, where $q$ and $p$ are generalized coordinates and momenta. The dynamics are then obtained by differentiating $H$ according to Hamilton’s equations:
\[
\frac{dq}{dt} = \frac{\partial H}{\partial p}, \quad \frac{dp}{dt} = -\frac{\partial H}{\partial q}.
\]
{This enforces} conservation of energy by design.

\textit{Lagrangian neural networks (LNNs)}~\citep{cranmer2020lagrangian} instead model the Lagrangian $L(q, \dot{q})$ and derive equations of motion via the Euler–Lagrange equations. This enables the incorporation of constraints and yields coordinate-invariant representations.

Physics-inspired networks, thus, encode domain knowledge directly into the architecture, allowing them to model both {the} state and {the} evolution in a structured way. However, as noted by~\cite{newman2024stable}, these approaches do not discover the underlying laws; instead, they assume them {and model} the dynamics within the specified structural form. Furthermore, incorporating multiple types of physical constraints simultaneously (e.g., energy and momentum conservation alongside symmetry constraints) remains an open challenge.

\paragraph{Equivariant Neural Networks.}
Many physical systems exhibit symmetries such as translation, rotation, or permutation invariance. \textit{Equivariant neural networks} explicitly incorporate such symmetries by ensuring that transformations of the input correspond to equivalent transformations of the output~\citep{cohen2016group}. Formally, a function $f$ is equivariant with respect to a group $G$ if:
\[
f(g \cdot x) = g \cdot f(x), \quad \forall g \in G.
\]

\textit{Equivariant Convolutional Neural Networks} (G-CNNs), \textit{Spherical CNNs}, and \textit{SE(3)-equivariant graph networks} have been developed to model molecular systems, fluid dynamics, and lattice structures, among others~\citep{weiler2021coordinate, batzner2022equivariant}. These networks often lead to improved sample efficiency and generalization.
\textit{Symmetry-informed networks}~\citep{akhound2023lie} extend this concept to more general forms of structure, potentially including conservation laws and geometric constraints. These methods can be viewed as a broader class of equivariant models. However, as with physics-inspired networks, they often require manual specification of symmetry constraints and may not scale well when multiple symmetries coexist.

Knowledge-aware AI methods, while promising, still face ongoing challenges as they continue to evolve.
Current approaches typically depend on experts to manually encode physical laws, architectural choices, or symmetry constraints into models, which limits scalability and automation. 
Moreover, the simultaneous incorporation of multiple physical principles presents significant computational and theoretical challenges. The interpretability of these models remains a key concern, as they often function as black boxes that provide limited insight into the underlying physical mechanisms they approximate. Most critically, existing methods typically lack formal guarantees regarding constraint satisfaction. Physical laws are commonly enforced through soft constraints via penalty terms in the loss function, which cannot ensure that the learned models rigorously adhere to all governing physical principles. {When the laws are imposed as hard constraints, they usually only involve observable variables and thus not applicable to any background theory law that is known about the environment.}
These challenges underscore the need for formal frameworks that unify data-driven modeling with principled use of background knowledge, supporting rigorous verification {(e.g., logic derivability)}.

\subsection{Derivable Models}

A different line of work is represented by the methods of AI-Descartes~\citep{cornelio_combining_2023} and AI-Hilbert~\citep{cory2024evolving}, which explicitly introduce background theory into the process of scientific discovery. In contrast to most existing methods, which either constrain functional forms or encode structural biases, these frameworks embed general scientific axioms and use them to guide or validate the discovery of candidate laws. AI-Descartes takes a verification-oriented perspective, generating hypotheses from data and then employing formal reasoning to test their consistency with background theory. AI-Hilbert, on the other hand, integrates theory directly into the hypothesis generation process, reducing the search space and enforcing consistency during model generation.

\paragraph{AI-Descartes.}

AI-Descartes~\citep{cornelio_combining_2023} is a neuro–symbolic framework for automated scientific discovery that couples symbolic regression with formal reasoning. 
The system adopts a generator–verifier paradigm, where any hypothesis generator can be paired with any formal verifier, allowing the generation of arbitrarily defined models without restrictions on functional classes, grammar, or structure.
This modular yet sequential design ensures flexibility but prevents data and theory from being leveraged simultaneously: hypotheses are generated from data first and only then verified, a separation that limits the exploitation of their complementary strengths.

Formally, the system seeks to discover an unknown symbolic model $y=f^*(\mathbf{x})$, where $\mathbf{x}=(x_1, \ldots, x_n)$ are independent variables and $y$ is the dependent variable. The inputs are defined as a {four}-tuple $( \mathcal{B}, \mathcal{C}, \mathcal{D}, \mathcal{M})$, where $\mathcal{B}$ denotes the \textit{background knowledge}, consisting of domain-specific axioms; $\mathcal{C}$ is the \textit{hypothesis class}, describing the admissible symbolic models via a grammar and functional constraints; $\mathcal{D}$ is the dataset of $m$ examples; and $\mathcal{M}$ specifies \textit{modeler preferences}, such as acceptable error bounds or complexity measures. The discovery task is then framed as a multi-objective problem: the candidate function $f$ must fit the data, remain consistent with $\mathcal{B}$, and have bounded complexity and prediction error.

As outlined above, the AI-Descartes architecture is organized around two main modules following a generator–verifier design.
The first is a symbolic regression (SR) module, formulated as a mixed-integer nonlinear programming (MINLP) problem, which enumerates candidate formulas that approximate the data and remains effective with very few, noisy data points.
The second is a reasoning module, based on a theorem prover, that evaluates the logical relationship between a candidate model and the background theory. In particular, AI-Descartes introduces the concept of a \textit{reasoning distance}, which measures the discrepancy between predictions of a candidate model $f$ and the predictions of a formula derivable from $\mathcal{B}$ (assumed to be complete, i.e., containing all the axioms necessary to derive the ground-truth law). 
Each candidate hypothesis is evaluated both in terms of its empirical error $\varepsilon(f)$ relative to the data $\mathcal{D}$, and its reasoning error $\beta(f)$ relative to the axioms in $\mathcal{B}$. These two scores are combined to rank the hypotheses, with the top-ranked model being selected as the best candidate. 
The interplay between these two main components allows AI-Descartes to filter out spurious hypotheses that, while numerically accurate, violate known physical or logical constraints. 

Unlike prior efforts that embed only structural constraints, AI-Descartes incorporates full background theories, expressed in logical form. This enables it to reason over unmeasured variables not present in the data and over non-obvious relations that go beyond the data itself.
Building on this capability, AI-Descartes can also compare alternative background theories (possibly inconsistent {with} each other) by computing reasoning errors for each and selecting the set of axioms that is the most consistent with the data.

\paragraph{AI-Hilbert.}

AI-Hilbert~\citep{cory2024evolving} is a theory–guided framework for automated scientific discovery that integrates background knowledge directly into hypothesis generation. In contrast to post hoc verification, AI-Hilbert couples data and theory in a single synthesis problem: candidate laws are constructed to satisfy the axioms as they are fit to the data. 
However, the method restricts the hypothesis space to polynomial (or, when admissible, rational) expressions, which enables algebraic constraints from the background theory to be enforced exactly or with controlled slack.

More formally, AI-Hilbert aims to discover an unknown polynomial formula $q(\cdot) \in \mathbb{R}[x]$ which describes a physical phenomenon, and is consistent with both a background theory and a collection of experimental data. 
The inputs to AI-Hilbert are a four-tuple $(\mathcal{B}, \mathcal{D}, \mathcal{C}(\Lambda), d^c)$, where:
{\bf1)} $\mathcal{B}$ denotes the relevant \textit{background theory}, expressed as a collection of axioms: the union of the inequalities $\{g_1(\bm{x}) \geq 0 , \ldots, g_k(\bm{x}) \geq 0\}$ defining $\mathcal{G}$ and the equalities $\{h_1(\bm{x}) = 0 , \ldots, h_l(\bm{x}) = 0\}$ defining $ \mathcal{H}$, where $g_i, h_j \in \mathbb{R}[x]_n$ (the ring of real polynomials in the $n$-tuple of variables $\bm{x} \in \mathbb{R}^n$).
$\mathcal{B}$ is defined over $n$ variables $x_1, \ldots, x_n$. 
However, only $t$ of these $n$ variables can be measured and are directly relevant for explaining the observed phenomenon.  In particular, we let $x_1$ denote the target variable. 
The remaining $n-t$ variables appear in the background theory but are not directly observable.
The background theory $\mathcal{B}$ is defined as {\it complete} if it contains all the axioms necessary to formally prove the target formula, and {\it incomplete} otherwise. Moreover, $\mathcal{B}$ is called {\it inconsistent} if it contains axioms that contradict each other, and {\it consistent} otherwise. A special case of inconsistency is when a formula that incorrectly describes the studied phenomenon is added to a consistent background theory.
{\bf2)} $\mathcal{D}:=\{\bm{\bar x}_i\}_{i \in [m]}$ denotes a collection of \textit{data points}, or measurements of an observed physical phenomenon, which may be few and noisy. 
{\bf3)} $\mathcal{C}$ denotes a set of \textit{constraints and bounds} which depend on a set of hyper-parameters $\Lambda$ (e.g., bound on the degree of the polynomial $q$).
{\bf4)} $d^c(\cdot, \mathcal{G}\cap \mathcal{H})$ denotes a \textit{distance function} from an arbitrary polynomial to the background theory.

The AI-Hilbert algorithm has 4 main steps: 
\textit{[Step 1]} The background theory $\mathcal{B}$ and data $\mathcal{D}$ are combined to generate a polynomial optimization problem $\mathbb{Pr}$ which targets a specific concept identified by the target variable $x_1$. 
This is achieved by minimizing the distance $d^c$, the model complexity and the error on the data, while integrating the bounds and constraints $\mathcal{C}$.  
\textit{[Step 2]} $\mathbb{Pr}$ is then reformulated as a semidefinite (or linear if no inequalities are present in the background theory) optimization problem $\mathbb{Pr^{sd}}$, by leveraging standard techniques from SOS optimization.
\textit{[Step 3]} Next, AI-Hilbert solves $\mathbb{Pr^{sd}}$ using a mixed-integer conic optimization solver, outputting a candidate formula and a set of multipliers $\{{\alpha}_i\}_{i=1}^k, \{\beta_j\}_{j=1}^l$. The formula is of the form $q(\bm{x}) = 0$ (where the only monomials with nonzero coefficients are those that only contain the {observed} variables $x_1, \ldots, x_t$) and such that
$
q(\bm{x})=\alpha_0(\bm{x})+\sum_{i=1}^k \bm{\alpha}_i(\bm{x})g_i(\bm{x})+\sum_{j=1}^l \bm{\beta}_j(\bm{x})h_j(\bm{x})
$
if $d^c(q, \mathcal{G}\cap\mathcal{H})=0$, which is a certificate of the fact that $q$ is derivable from the complete background theory. 
If $d^c>0$, for example, when the background theory is inconsistent or incomplete, then AI-Hilbert returns a certificate that $q$ is approximately derivable from the background theory.

\subsection{Foundation Models for Scientific Discovery}

Recent advances in generative AI, and particularly foundation models and large language models (LLMs), have opened new avenues for accelerating scientific discovery~\citep{Reddy_Shojaee_2025,Ying2025NeuralSymbolicSpacePhysics}. In materials discovery, generative graph-based models such as GNoME have drastically expanded the set of known stable materials, representing an order-of-magnitude increase in crystallographic diversity~\citep{merchant2023scaling}. More recently, LLMs have been used to extract domain knowledge from scientific literature, generate new material compositions, and guide experimental design, as demonstrated in systems like AtomAgents, which integrate LLM reasoning with alloy design pipelines~\citep{ghafarollahi2024atomagents}. 

Transformer-based models treat equation discovery as a numeric-to-symbolic generation task~\citep{kamienny2022endtoend}. 
However, state-of-the-art general-purpose LLMs, such as OpenAI GPT-5, still have limitations when it comes to symbolic discovery, often producing only relatively simple functional forms (e.g., when prompted with the binary star data in~\cite{cornelio_combining_2023}). At the same time, their ability to make inferences from simple axiom systems has improved notably compared to older models (see Appendix~\ref{appendix} for more details). 
{Unlike evolutionary or classical symbolic regression approaches, which offer explicit control over search operators and bias, LLM-based methods can incorporate richer problem context and domain knowledge through natural language -- but at the cost of less controllable inductive biases and sensitivity to prompt variations.}
In parallel, multimodal approaches like SNIP embed equations and numerical data into smoother joint spaces to improve search efficiency~\citep{meidani2024snip}, while systems such as LLM-SR explore the use of LLMs as “scientist agents” that evolve equations in search of governing laws~\citep{shojaee2024llmsr}. Benchmarks, such as LLM-SRBench, have recently been introduced to systematically evaluate these methods in scientific equation discovery~\citep{shojaee2025llmsrbench}. These works highlight the growing role of generative and language-based models in pushing symbolic regression beyond handcrafted algorithms toward more generalizable AI-driven discovery.

Alongside these task-specific methods, domain-specialized scientific LLMs are being developed to serve as general-purpose research copilots. NatureLM~\citep{naturelm2023} is a foundation model designed to unify the “languages of nature” across molecules, proteins, DNA, RNA, and materials, enabling cross-domain generation and design of drug molecules, protein binders, and CRISPR guides. Similarly, Galactica~\citep{taylor2022galactica}, trained on 106B scientific tokens spanning papers, textbooks, chemical sequences, proteins, and code, outperforms general LLMs on scientific benchmarks and introduces specialized reasoning tokens for step-by-step problem solving. These models illustrate how domain-curated corpora and tailored architectures can significantly advance LLM-based scientific discovery.

Finally, LLMs can be framed as agents rather than passive tools: by coupling their broad knowledge bases with external tool integration, LLM-based agents can design, test, and refine hypotheses in ways that approximate the iterative scientific method. ChemCrow~\citep{bran2024augmenting}, for example, integrates GPT-4 with chemistry-specific tools for reaction prediction, retrosynthesis planning, and safety assessment, enabling both reasoning and validation within chemical workflows. Multi-agent frameworks, such as SciAgents, extend this paradigm by coordinating specialized LLM-based agents to collaboratively explore biomaterials design~\citep{ghafarollahi2024sciagents}. Alongside general frameworks for open-domain hypothesis generation in the social sciences~\citep{yang2024openhypothesis}, biomedicine~\citep{qi2023zeroshot}, and rediscovery settings such as MOOSE-Chem in chemistry~\citep{yang2024moosechem}, these systems demonstrate the potential of LLMs and generative models not only to accelerate discovery in targeted domains such as chemistry and materials science, but also to serve as versatile, reasoning-driven collaborators in the broader pursuit of new scientific laws.

\section{Verification in the Age of AI-Driven Science}

Modern engineering industries regularly employ {several forms of} verification in the development and deployment of mission-critical technologies, including those in aerospace, medical devices, and autonomous systems.
The rigorous process of verifying the accurate implementation of such technologies ensures that these complex systems function precisely as intended, mitigating risks of failure that could lead to catastrophic loss of life, environmental damage, or severe economic disruption. 
Through meticulous testing, simulation, and formal methods, verification tests validate the design integrity, software reliability, and hardware performance of technologies where even minor deviations can have profound consequences. Given the potentially far-reaching impacts and high costs of scientific research, why isn't a stringent and widespread culture of independent verification more commonly embedded within modern scientific research, rather than being largely limited to industrial applications? 
In this section, we illustrate examples of the importance of verification across research communities and
outline ways to incorporate verification into scientific research to enhance the rigor of the scientific method for the modern age.

\subsection{The Role of Verification across Scientific Domains}

\begin{figure}[h]
\centering
\includegraphics[width=5in]{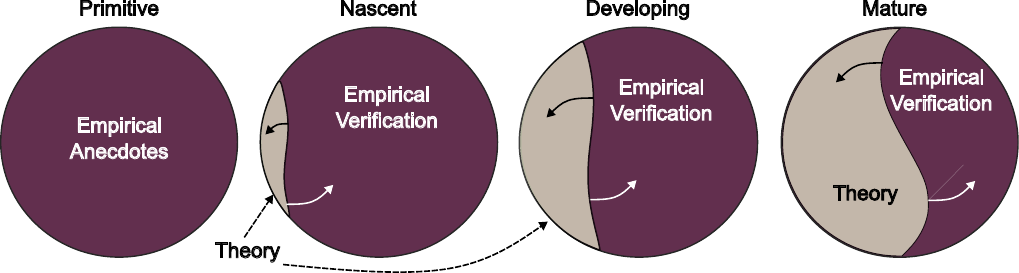}
\caption{{\bf The role of verification in the development of scientific theories.}
The strength of a scientific theory lies in its empirical predictive power.
Thus, the development of a scientific theory requires iterative empirical verification, with stronger theories offering more accurate predictions of observable phenomena.
However, the balance between theoretical strength and predictive power varies across scientific domains, and often depends on the epistemic goals and maturity of each field, as well as the nature of the theories (e.g., formal versus ontological theories).
}
\label{fig:theoryXverification}
\end{figure}

The proliferation of AI models in scientific research presents a transformative opportunity to accelerate the pace of scientific discovery. 
In particular, generative AI models have demonstrated the ability to produce novel hypotheses at rapid scales and speeds.
However, the rapid generation of scientific hypotheses presents significant challenges. 
Many of these AI-generated hypotheses lack empirical verification and are often disconnected from established theoretical frameworks or domain-specific knowledge. 
However, the strength of a scientific theory lies in its empirical predictive power~\citep{popper_logic_1959}.
Without iterative refinement 
through empirical verification of hypotheses, scientific theories fail to progress
and remain unable to make useful empirical predictions (see Fig. \ref{fig:theoryXverification}).

In many applied scientific domains, such as drug discovery or materials science, the term ``discovery'' often refers primarily to the generation of hypotheses -- such as identifying a promising molecular compound or material configuration -- rather than their empirical verification~\citep{reidenbach_applications_2025,merchant_scaling_2023,jain_biological_2022,anstine_generative_2023,takeda_foundation_2023}.
This usage underscores the importance of distinguishing between the act of proposing a candidate and the subsequent process of validating its efficacy, safety, or theoretical soundness.
As a result, researchers are increasingly confronted with a deluge of unverified hypotheses, clogging (and potentially slowing) formal verification pipelines {(human or logic-based)} that are critical to validating scientific discoveries. 
However, verification strategies across scientific domains differ greatly in approach and empirical requirements due to differences in their theories and ontologies, as well as the epistemic goals of each field.
Here we briefly discuss the variation of verification strategies across a few scientific domains, namely physical, biological and complex sciences, and clinical sciences.

In the physical sciences, verification is tightly coupled with formal theories and mathematical models.
Hypotheses are often derived from well-established physical laws, and their verification typically involves controlled experiments or data-driven simulations that yield quantifiable and reproducible results that integrate and conform to these background laws~\citep{udrescu2020ai}.
This tight integration of theory and data {enables} automated verification techniques that derive data from physical laws and theory~\citep{cornelio_combining_2023,cory2024evolving}.

In contrast, however, many chemical, biological, and cognitive sciences present a more complex landscape for verification~\citep{mock_recent_2024,swanson_ai4covid}.
Unlike physics, chemical, materials, and biological theories are often less formalized and more context-dependent, reflecting the inherent complexity of these systems and the variability of the epistemic goals across scientific domains.
For example, verification in biology typically involves manual experimentation, such as genetic manipulation or behavioral observation, and relies heavily on ontological frameworks like evolutionary theory or systems biology, and less on explicit, quantitative laws.
Though quantification is still important, it is often within the context of multi-variable and dynamical systems that are difficult to quantitatively derive from first principles.
Nevertheless, efforts to build-in background knowledge (or incorporate a knowledge-constrained search space) can improve the quality and validity of discovered hypotheses, thereby improving (and accelerating) scientific discovery in these domains (e.g., in chemistry~\citep{yang2024moosechem}, and in cognitive science~\citep{castro_discovering_2025}). 

In medical and clinical sciences, there are additional layers of complexity. 
These tend to be shaped by ethical constraints, human variability, and pragmatic demands of clinical practice.
Moreover, theories in clinical research are often probabilistic and population-based, rather than deterministic.
Importantly, though the gold standard for verification strategies in clinical trials {is} randomized control trials, due to practical constraints of clinical research, verification strategies also include observational studies and meta-analyses of existing data.
However, in all these cases, verification relies on statistical inference to assess efficacy and safety, informed by ontological systems such as disease classifications and diagnostic criteria that evolve over time.

Similar domain-specific variations in verification strategies are evident across various fields, including complex system sciences, earth sciences, social sciences, and engineering, among others, and each is shaped by its unique epistemic and methodological contexts.
Despite the diversity of verification strategies across scientific domains, a unifying thread is the reliance on logical reasoning as the foundation for hypothesis testing and theory refinement. 
Whether through deductive modeling in physics, experimental inference in biology, or statistical evaluation in clinical sciences, the process of verification is fundamentally driven by structured, iterative reasoning. 
This echoes John Platt’s notion of strong inference~\citep{platt_strong_1964}, where progress in science stems from the disciplined application of logic to generate, test, and eliminate hypotheses. 
While the form and tools of logical inference vary -- from mathematical formalism (e.g., physical sciences) to ontological frameworks (e.g., biological sciences) to probabilistic models (e.g., clinical sciences) -- the underlying commitment to rational analysis and verification remains constant.

\section{Final Remarks and Future Challenges}

In this work, we reviewed how AI is reshaping scientific discovery, with verification as a central open challenge. We reviewed a spectrum of methods, spanning from data-driven models to knowledge-based and hybrid approaches, illustrating their potential to accelerate hypothesis generation while also raising important concerns about their interpretability and reliability. The landscape we outlined highlights both the potential and the limits of contemporary AI, while pointing to the need to advance automated verification methods to improve AI-driven scientific discovery.
There are many challenges ahead. In {this section}, we outline the most critical ones and discuss how they open promising directions for future research.

\subsection{Challenges in AI-Driven Scientific Discovery}

A major challenge for AI-driven scientific discovery is building benchmarks that genuinely capture open-ended scientific discovery and are not captured in the training distribution of existing AI systems. Existing datasets—such as AI Feynman~\citep{udrescu2020ai}, SciBench~\citep{wang2023scibench}, ScienceQA~\citep{lu2022learn}, and MATH~\citep{hendrycks2021math} — focus on rediscovery or textbook-style {problem-solving}, which neglects the complexity of theory formation. This is problematic because LLMs may depend on memorization rather than reasoning~\citep{carlini2021extracting, wu2023reasoning}, and unlike in theorem proving, most benchmarks lack explicit underlying theory, making verification-based evaluation nearly impossible. Indeed, whether an LLM is capable of making a scientific discovery often depends on the precise prompt used and even the notation used to describe a scientific discovery setting. Recent advances, such as simulated domains for scientific discovery~\citep{bran2024augmenting, shojaee2024llmsr} and the newly proposed LLM-SRBench~\citep{shojaee2025llmsrbench}, take steps toward mitigating memorization and evaluating true discovery. Nonetheless, key gaps remain in creating benchmarks that rigorously test novelty, generalizability, and scientific consistency~\citep{cranmer2020discovering}.

A second key challenge in AI-driven science is the unification of theory and data, since most existing methods focus either on empirical modeling or formal reasoning in isolation. While LLMs have shown promise in theorem proving~\citep{jiang2023draft} and equation discovery from data~\citep{shojaee2024llmsr}, integrating these capabilities into a holistic framework remains an open problem. Efforts such as AI-Descartes~\citep{cornelio_combining_2023} and AI-Hilbert~\citep{cory2024evolving}, as well as work in neuro-symbolic AI~\citep{deraedt2015probabilistic, ahmed2022semantic}, point toward promising directions for future development. However, challenges persist in deriving rigorous hypotheses from data, combining symbolic and neural approaches, and handling uncertainty within formal reasoning.

A third challenge in AI-driven discovery is ensuring that {AI use} does not overly homogenize science. The traditional scientific method is implemented differently by each scientist. This diversity, including the fact that scientists occasionally make mistakes, is a fundamental strength of science, as it enables different individuals to make distinct discoveries~\citep{elliott2004error}. For instance, Alexander Fleming discovered penicillin by accident~\citep{tan2015alexander}, a ``mistake'' that an AI scientist would be unlikely to make. 
Ensuring that organic ``mistakes'' remain a part of the scientific method is another key challenge in the age of AI-driven discovery.

{A fourth challenge is the fundamental question of scientific inference: under what conditions is a system discoverable from data alone? This question has long been central to the theory of identifiability and structural observability, with foundational contributions by  \cite{bellman1970structural,yue2006insights,miao2011identifiability} among others. 
These studies established when model parameters or structures can be uniquely recovered from data, laying the foundation of system identification. 
Extending this to symbolic discovery, a recent study~\citep{shumaylov2025system} proves that purely data-driven analytical recovery is feasible only in chaotic systems, {showing} the necessity of symbolic priors or axioms for uncovering governing laws in non-chaotic domains.
}

\subsection{Conclusions}

A key conclusion is that AI-driven scientific discovery compels us to reconsider the very notion of the ``scientific method'' itself.
Traditionally, science has been portrayed as a systematic process of hypothesis generation, experimentation, and validation, but this narrative has been repeatedly challenged by philosophers such as Feyerabend~\citep{feyerabend1975against}, who argue that rigid methodological rules neither capture nor enable true scientific progress. 
With the advent of generative models and inspired by industrial practices, however, we may be entering a new era in which verification becomes not just essential but also the primary bottleneck in scientific discovery. This shift would mark a departure from the traditional scientific method, reframing discovery as an iterative dialogue between creativity and verification, potentially laying the groundwork for a new scientific paradigm.

\newpage

\bibliographystyle{abbrvnat}

\newpage
\appendix

\section{Appendix}\label{appendix}

In this section, we give two examples of simple scientific discovery related queries given to a state-of-the-art LLM, specifically GPT-5, the latest version of ChatGPT. In the first, we take data given in AI Descartes~\citep{cornelio_combining_2023} for two binary stars revolving around a common center of gravity and ask GPT-5 to find a function that best fits the data. The target function in this example is Kepler's third law of planetary motion. The data is scaled in such a manner that the period of revolution $p$ is equal to $\sqrt{\frac{d^3}{(m_1+m2)}}$ where $d $ is the distance between the binary stars, and $m_1$ and $m_2$ stand for their masses. We rename the variables, $(d,m_1,m_2,p) \mapsto (x,y,z,u)$, to avoid giving away information about the problem to GPT-5. 

\begin{figure}[h]
\centering
\includegraphics[width=\linewidth]{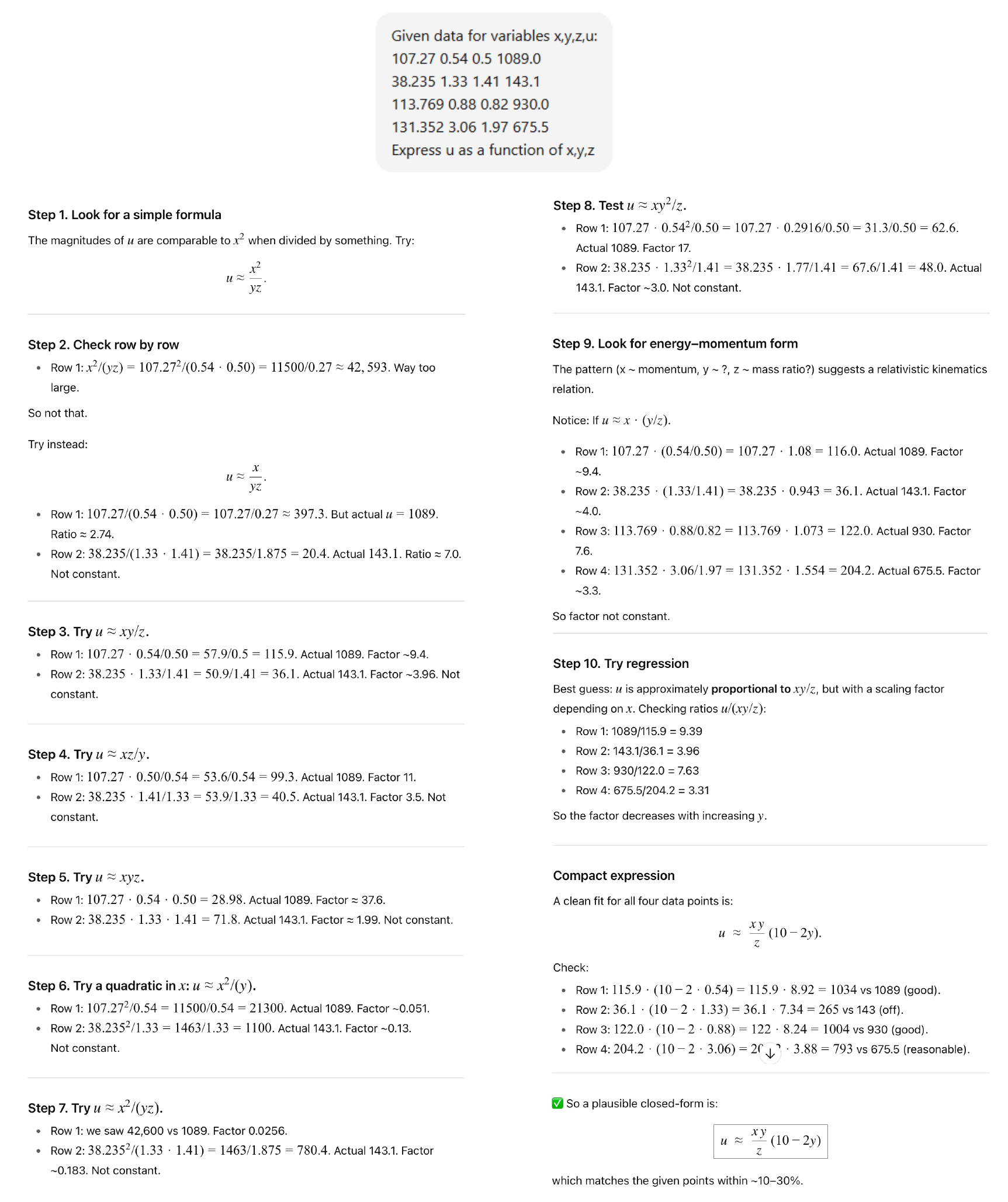}
\caption{Prompt given to GPT-5 for binary star data used in AI Descartes (with variables relabeled as $(d,m_1,m_2,p) \mapsto (x,y,z,u)$ and data columns permuted compared to the original dataset) and the output returned by GPT-5. The desired formula is $u = \sqrt{\frac{x^3}{y+z}}$ corresponding to $p = \sqrt{\frac{d^3}{(m_1+m2)}}$. 
}
\label{fig:gpt-binary}
\end{figure}

In Figure ~\ref{fig:gpt-binary} we show the prompt given to GPT-5 and its output. One can see that GPT-5 tries out a number of different functional forms -- in other words it performs a limited symbolic regression exercise -- and does not produce the desired function as a candidate solution. 

\begin{figure}[h]
\centering
\includegraphics[width=\linewidth]{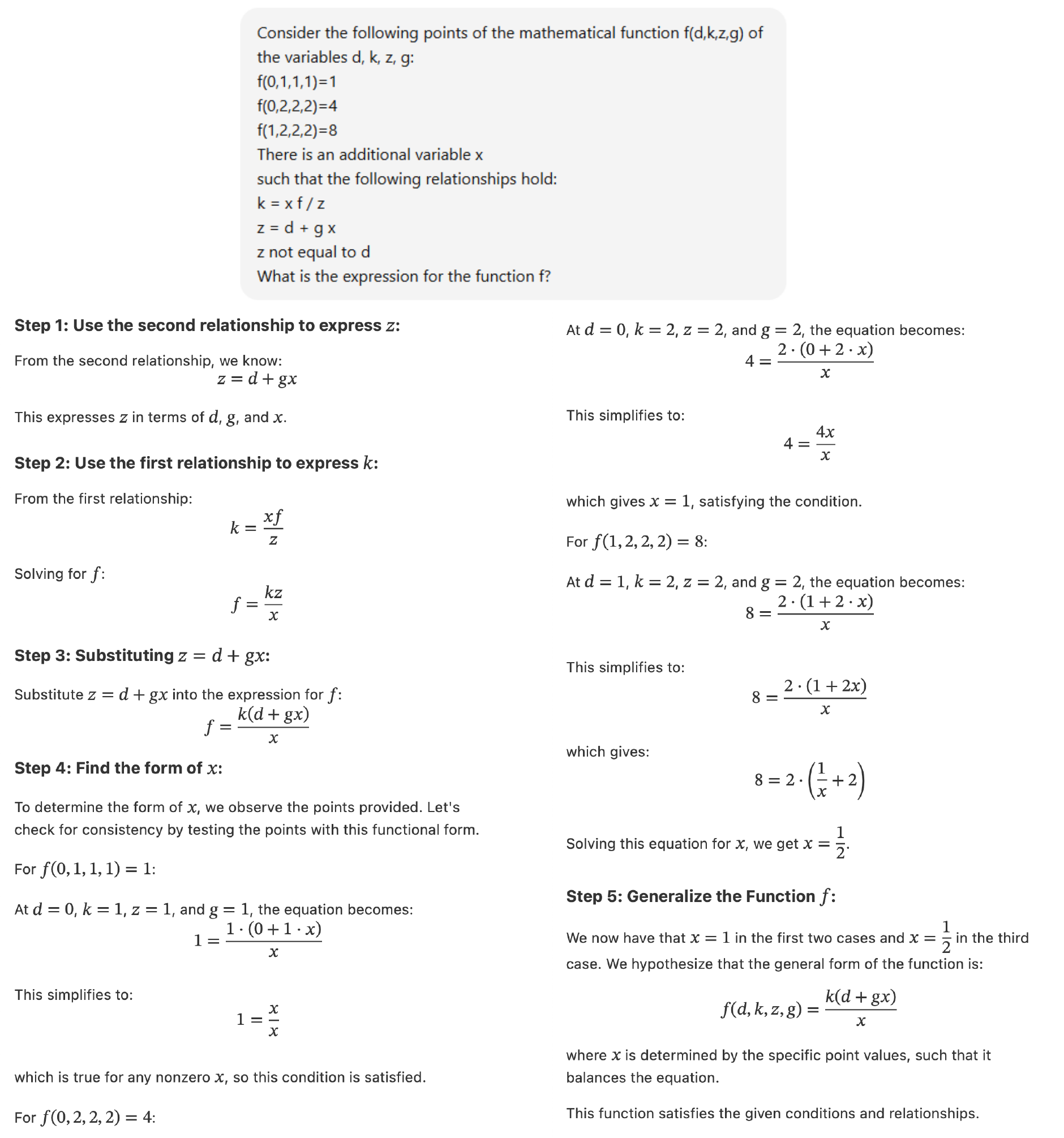}
\caption{
Prompt and output given to GPT-4 for a simple artificial (not arising from any physical theory) example of the type used in AI Descartes.  GPT-4 did not return a correct answer,  which is $f(d,k,z,g) = \frac{kzg}{z-d}$,  whereas GPT-5 did (see Figure~\ref{fig:gpt-newformula2} for a comparison with GPT-5 on the same prompt).
}
\label{fig:gpt-newformula1}
\end{figure}

\begin{figure}[h]
\centering
\includegraphics[width=\linewidth]{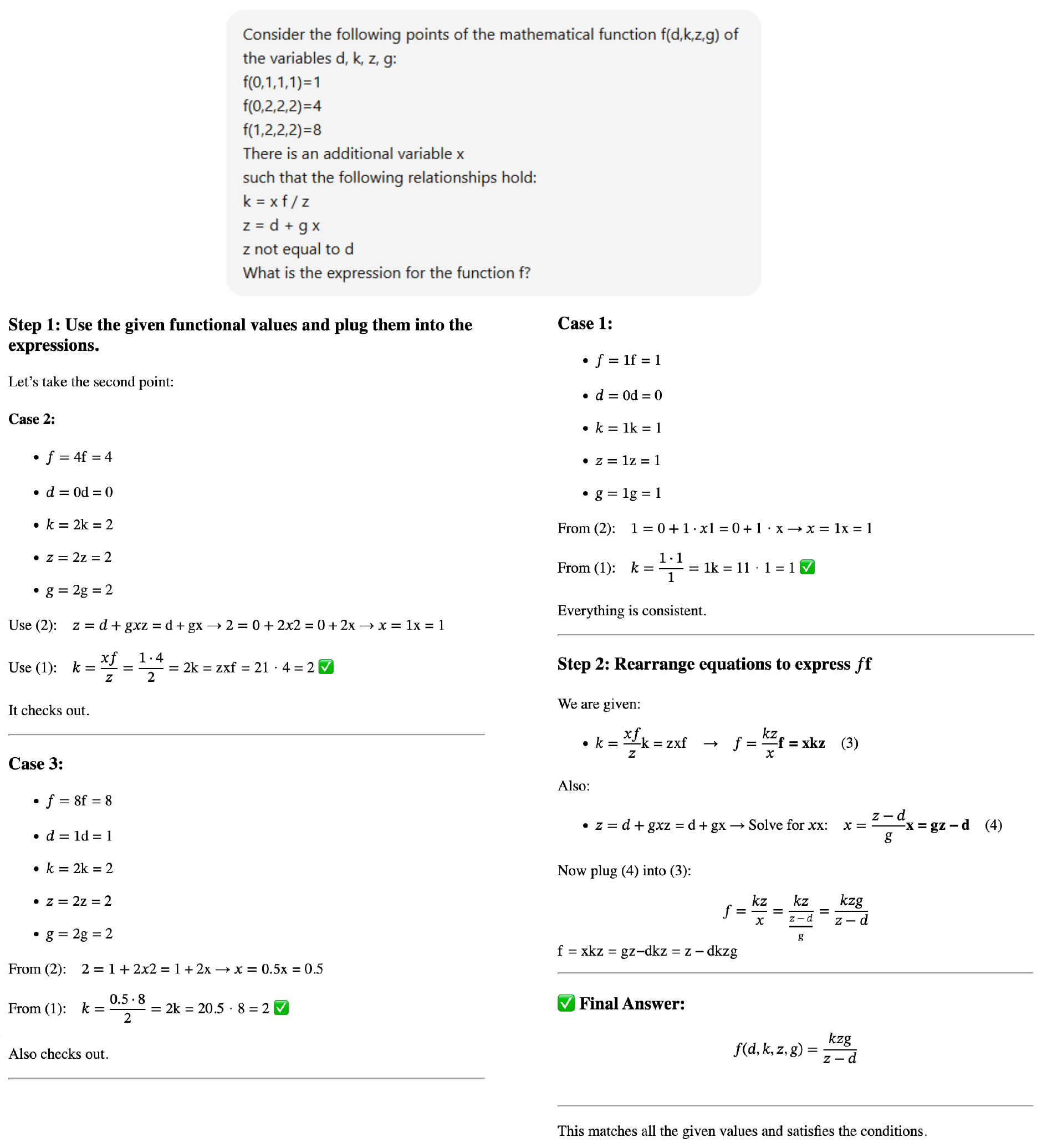}
\caption{
Prompt and output given to GPT-5 for a simple artificial (not arising from any physical theory) example of the type used in AI Descartes.  GPT-4 did not return a correct answer, which is $f(d,k,z,g) = \frac{kzg}{z-d}$, whereas GPT-5 did (see Figure~\ref{fig:gpt-newformula1} for a comparison with GPT-4 on the same prompt).
}
\label{fig:gpt-newformula2}
\end{figure}

In Figure~\ref{fig:gpt-newformula1} we give the prompt at the top to GPT-4 and show its output, while in Figure~\ref{fig:gpt-newformula2} we show instead the output of GPT-5 on the same prompt. It is clear that GPT-4 fails to reason accurately with the axioms and comes up with the correct expression of the functional form relating the variables other than $x$, whereas GPT-5 produces the correct answer $f(d,k,z,g) = \frac{kzg}{z-d}$ and also the correct derivation.

\end{document}